\documentclass[journel]{IEEEtran}
\IEEEoverridecommandlockouts
\usepackage{cite}
\usepackage{amsmath,amssymb,amsfonts}
\usepackage{algorithmic}
\usepackage{graphicx}
\usepackage{textcomp}
\def\BibTeX{{\rm B\kern-.05em{\sc i\kern-.025em b}\kern-.08em
    T\kern-.1667em\lower.7ex\hbox{E}\kern-.125emX}}
\usepackage{multirow}
\usepackage{adjustbox}
\usepackage{amsmath}
\usepackage{amssymb}
\usepackage{subfigure}
\usepackage{booktabs}
\usepackage{comment}
\usepackage[normalem]{ulem}
\usepackage[ruled,linesnumbered]{algorithm2e}
\usepackage[dvipsnames]{xcolor}

\newtheorem{definition}{Definition}[section]
\usepackage{amssymb}

\newif\ifmodify
\modifytrue 

\ifmodify
\newcommand{\cross}[1]{\textcolor{red}{\sout{#1}}}

\else
\newcommand{\cross}[1]{}

\fi

\newif\ifmaximal
\maximaltrue 

\title{Load-balanced Gather-scatter Patterns for Sparse Deep Neural Networks}
\author{
    \IEEEauthorblockN{Fei Sun\textsuperscript{1}, Minghai Qin\textsuperscript{2}, Tianyun Zhang\textsuperscript{3}, Xiaolong Ma\textsuperscript{4}, Haoran Li\textsuperscript{1}, \\Junwen Luo\textsuperscript{1}, Zihao Zhao\textsuperscript{5}, Yen-Kuang Chen\textsuperscript{1}, Yuan Xie\textsuperscript{1}
    \thanks{\textsuperscript{2} Work done while working at Alibaba DAMO Academy. }
    \thanks{\textsuperscript{3,4,5} Work done while interning at Alibaba DAMO Academy.}\\}
    \IEEEauthorblockA{\textsuperscript{1}Alibaba DAMO Academy \\
    \textsuperscript{2} Western Digital Research \\
    \textsuperscript{3} Cleveland State University \\
    \textsuperscript{4} Northeastern Unviersity \\
    \textsuperscript{5} Fudan University
    }
}

\begin{document}

\maketitle

\begin{abstract}
Deep neural networks (DNNs) have been proven to be effective in solving many real-life problems, but 
its high computation cost prohibits those models from being deployed to edge devices. 
Pruning\ifmaximal, as a method to introduce zeros to model weights, \fi 
~has shown to be an effective method to provide good trade-offs between model accuracy and computation efficiency\ifmaximal, and is a widely-used method to generate compressed models\fi. 
However, the granularity of pruning makes important trade-offs. At the same sparsity level, a coarse-grained structured sparse pattern is more efficient on conventional hardware but results in worse accuracy, while a fine-grained unstructured sparse pattern can achieve better accuracy but is inefficient on existing hardware.

On the other hand, some modern processors are equipped with fast on-chip scratchpad memories and gather/scatter engines that perform indirect load and store operations on such memories. In this work, we propose a set of novel sparse patterns, named {\em gather-scatter (GS)} patterns, to utilize the scratchpad memories and gather/scatter engines to speed up neural network inferences. 
\ifmaximal Correspondingly, we present a compact sparse format. \fi
The proposed set of sparse patterns, along with a novel pruning methodology, address the load imbalance issue and result in models with quality close to unstructured sparse models and computation efficiency close to structured sparse models.
Our experiments show that GS patterns consistently make better trade-offs between accuracy and computation efficiency
compared to conventional structured sparse patterns. GS patterns can reduce the runtime of the DNN components by two to three times  at the same accuracy levels. This is confirmed on three different deep learning tasks and popular models, namely, GNMT for machine translation, ResNet50 for image recognition, and Japser for acoustic speech recognition. 
\end{abstract}
\section{Introduction}

With the advancement of the deep learning algorithms, many applications have seen significant accuracy improvement, such as image classification~\cite{krizhevsky2012imagenet}, object detection~\cite{ren2015faster}, semantic segmentation~\cite{long2015fully}, and natural language processing (NLP)~\cite{devlin2019bert}. The eager to obtain higher accuracy
pushes the model computation to the limit of the available hardware. 
\ifmaximal For example, {the GPT-3 model} consists 175 billion parameters, and training it requires a super-computer~\cite{brown2020language}. \fi

On mobile and edge devices, the desire for efficient models is even stronger. With limited memory and computation resources, model sizes are usually constrained to sacrifice model quality. Among various approaches to limit model sizes~\cite{han2015deep,park2017weighted}, weight pruning has shown to be an effective mechanism~\cite{han2015deep,mao2017exploring,wen2016learning,he2017channel}. 
\ifmaximal
Weight pruning introduces zeros to the weight matrices to create sparse models. Those zeros no longer need to be stored in memory during inferences, and thus reducing both memory consumption and computation latency. 
\fi

Many pruning techniques have been researched to trade-off model quality and computation efficiency~\cite{wen2016learning,han2015deep,zhang2018systematic,ma2020image,liu2018darts,lin2019towards,verelst2020dynamic}. Irregular sparsity~\cite{han2015deep}, where each value in the weight matrices is individually determined to be zero or not by the pruning algorithm, usually results to better model quality.
However, it is less hardware friendly because of its irregular memory access patterns. Structured sparsity~\cite{wen2016learning}, where a group of values in the weight matrices are determined to be zero or not as a unit, is more hardware efficient, but is less accurate at the same sparsity level as the irregular sparsity. 

{Efficient fine-grained sparsity is proven effective in speeding up DNN computation on A100 GPUs~\cite{nvidia2020a100} with the introduction of the sparse tensor cores.  However, it is not yet explored on CPUs and DSPs. Yet many of them have already equipped with the necessary hardware. 
DSPs like Tensilica~\cite{Tensilica}, CEVA~\cite{CEVA}, and ARC~\cite{ARC} may optionally include one or multiple on-chip static random-access memories (SRAMs) called tightly coupled memories (TCMs). }
Such memories are divided into banks and sub-banks, and each sub-bank is separately addressable. 
A gather/scatter engine can load or store different data {from or} to different sub-banks in parallel. The TCM and gather/scatter engine are widely used in many classical applications 
but it is not yet explored to use them to speed up sparse model inference. One challenge of using gather/scatter engines on sparse models is to remove the load imbalance of the irregular data accesses.

In this work, we note that load imbalance is mainly due to the non-zero weights that point to some sub-banks in TCM more than others. If a sparse pattern can balance the locations of non-zero weights, we can eliminate bank conflicts in the gather-scatter engine.
While being much more hardware friendly, the GS patterns achieve the same accuracy as irregular sparse pattern. Experimental results of three different deep learning model architectures and tasks have confirmed the improvement of accuracy-latency trade-off by the proposed GS sparse pattern.


The contribution of this paper is summarized as follows.
\begin{itemize}
    \item We propose a set of sparse patterns, named {\em gather-scatter (GS)} patterns, for weight matrices. They can fully take advantage of the computation characteristics of gather-scatter engines and minimize the bank access conflicts for the given sparsity levels. Compact sparse formats are also proposed for the corresponding GS patterns.

    \item With the proposed sparse patterns, we demonstrate that GS patterns achieve the same accuracy as the irregular pattern, which is \ifmaximal the most flexible and \fi theoretically the upper bound on the accuracy. This observation is confirmed on three different deep learning tasks and model architectures, namely, GNMT (LSTMs) for machine translation, ResNet50 (2-D CNNs) for image recognition, and Jasper (1-D CNNs) for acoustic speech recognition.
    
    \item With the same accuracy, we demonstrate that the proposed GS sparse patterns achieve $2-3\times$ runtime latency reduction (in clock cycles during inferences, simulated by Gem5) compared to the block sparse pattern, which is one of the most hardware-friendly structured patterns.

\end{itemize}

\section{Motivation}

Models with structured sparse patterns are more computational efficient than irregular sparse patterns~\cite{wen2016learning}. However, in order to achieve the same model quality, models pruned with structured sparse patterns need to rest at a much lower sparsity level. This significantly reduces their runtime speedup potential. 

\begin{figure}[tbp]
\centering
 \includegraphics[width=0.9\linewidth]{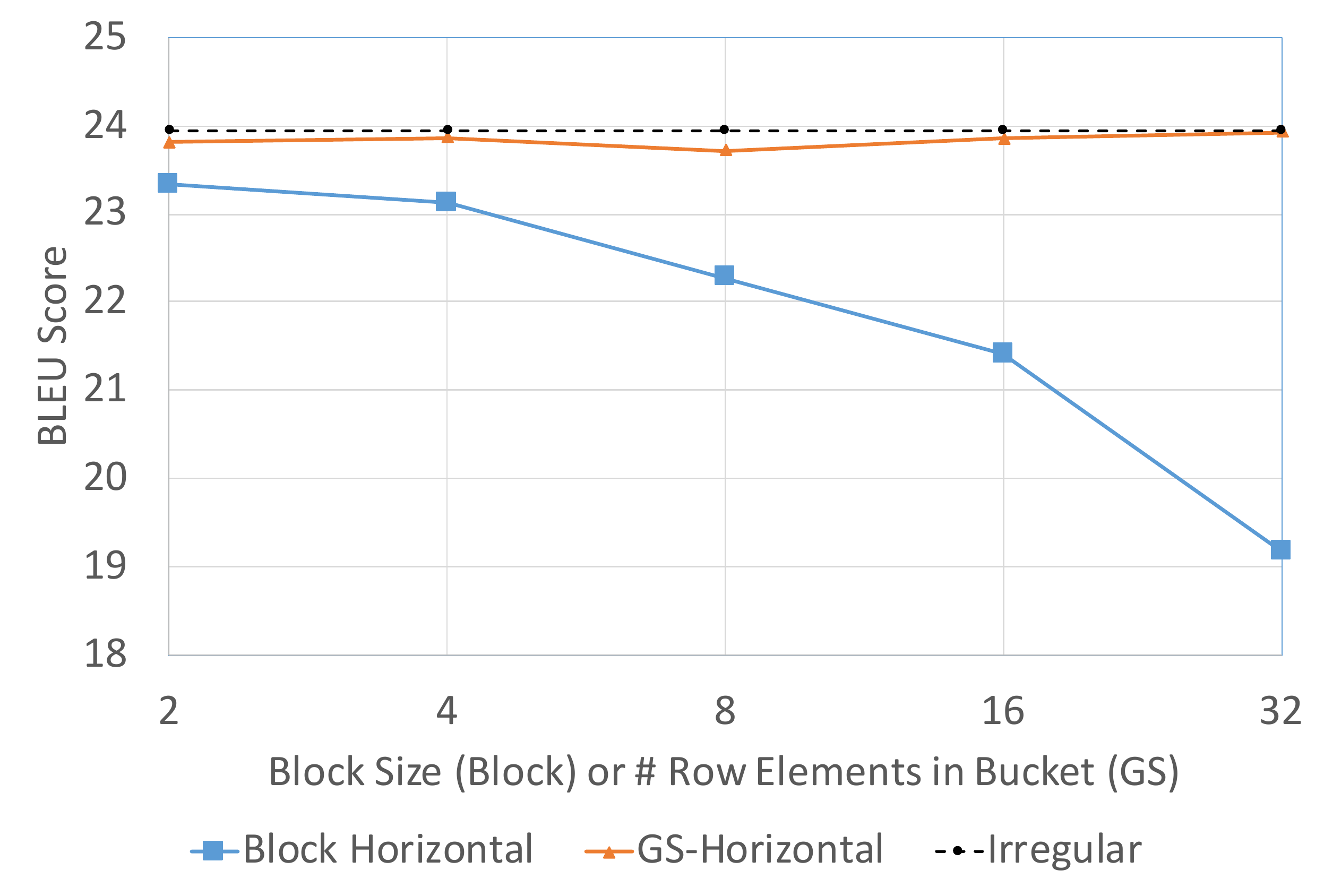}

  \caption{\small BLEU scores of the GNMT model with block horizontal sparse patterns and gather-scatter horizontal sparse patterns. All models are at 90\% weight sparsity. X-axis {is} the length of the block or the number of sub-banks for GS horizontal patterns. Y-axis {shows} the BLEU scores evaluated on WMT De $\rightarrow$ En test set. }
  \label{fig:blk_vs_gs_0.9}
\vspace{-0.1in}
\end{figure}

To illustrate this limitation, we have performed a series of experiments on the GNMT model~\cite{wu2016google}, as shown in Figure~\ref{fig:blk_vs_gs_0.9}. The experiment setup is described in {Supplementary material}. All models are pruned to 90\% sparsity. The block structured sparse patterns set $2$, $4$, $8$, $16$, or $32$ consecutive weight values along the reduction dimension to be zero or non-zero as a unit.  
\ifmaximal
We call this block sparse pattern as {\em block horizontal} and the consecutive weight values in the other perpendicular dimension is called {\em block vertical}. 
\fi
As shown in the blue line in Figure~\ref{fig:blk_vs_gs_0.9}, the BLEU scores of the models pruned using larger block sizes are materially decreased from the models with smaller block sizes. Since the block size usually matches the SIMD width of the processor, higher efficiency is obtained with larger block sizes. However, to compensate the decreased BLEU scores, lower sparsity levels are resorted and thus reducing the speedup.

In this work, we propose GS patterns that can efficiently utilize the SIMD instructions as the block sparse pattern, but the accuracies with larger  GS pattern sizes do not degrade from the irregular pattern, as shown in the top two curves in Figure~\ref{fig:blk_vs_gs_0.9}. In this way, higher sparsity levels can be achieved to boost the speedup.



\section{Background on Gather/Scatter Engine}
\label{sec:processor}

\begin{figure}[t]
\centering
  \begin{center}
  \begin{tabular}[t]{ @{}c@{} }
  \includegraphics[width=0.98\linewidth]{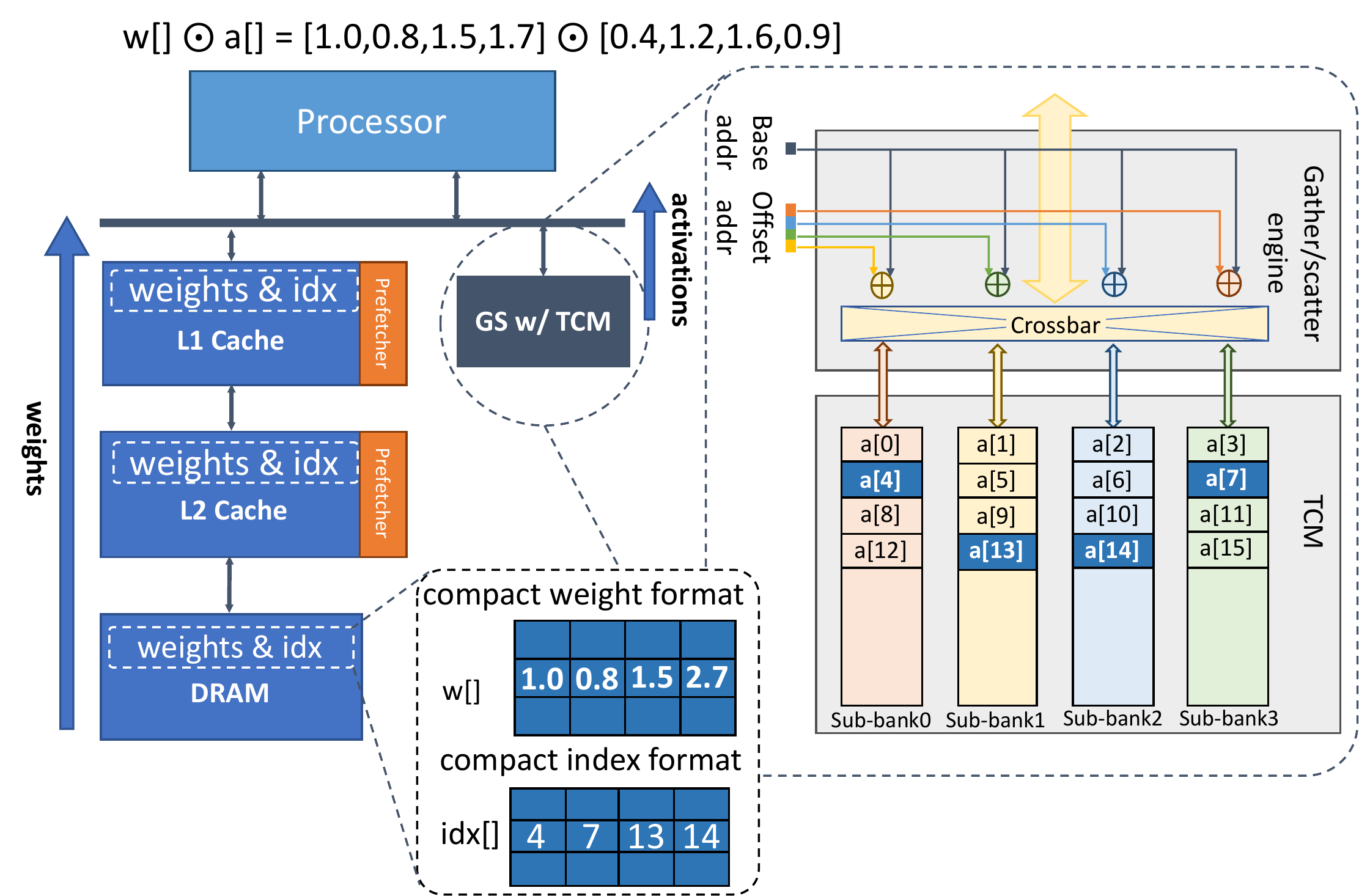} \\
  {\small (a)} \\
  \includegraphics[width=0.98\linewidth]{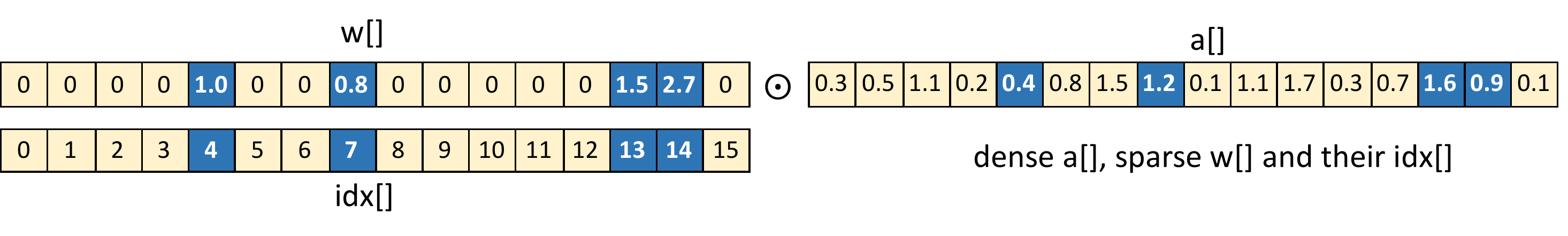} \\
  {\small (b)} \\
  \end{tabular}
  \end{center}
  \caption{\small Illustration of the processor memory hierarchy with TCM and gather/scatter engine. It shows an element-wise product of sparse weights w[] and dense activations a[]. The sparse weight vector is stored in a compact representation (w[] and idx[]) in the memory hierarchy and loaded to the processor. The dense activation vector is sequentially stored across 4 sub-banks in the TCM.}
  \label{fig:overall_arch}
\end{figure} 

In deep learning models, weights can be pruned to be sparse and saved in the compressed format beforehand. 
The sparse computation in deep learning models on processors often refer to  sparse weights multiplying with the dense activations. 
In such computation, the sparse weights are accessed sequentially, as it is already saved in the compressed format. But the indices of the weights are used to reference the matching activation values, resulting to random accesses to the activations. 
Because of the random accesses, the local data reuse on  sparse computation is significantly reduced compared with the dense counterpart. 
\ifmaximal
Techniques such as tiling at the register file or L1 cache level cannot be effectively exploited. 
Due to this characteristic of sparse computation, it 
\fi
is more efficient to stream the sparse weights into the processors from the lower level memory hierarchy, while the randomly accessed activations are kept close to the processors to reduce the access latency. This is illustrated in the data flow in Figure~\ref{fig:overall_arch} (a).

On the other hand, many modern processors, especially data processing processors~\cite{Tensilica, CEVA, ARC}, can set up large TCMs on-chip and use gather/scatter engines to access such memories, as shown in Figure~\ref{fig:overall_arch}~(a)~\footnote{The gather/scatter instructions in x86 architecture do not use TCM and cannot achieve the same performance.}. The TCMs are composed of multiple banks and sub-banks, with data elements interleaved in the sub-banks at low order bits. As each sub-bank is individually addressable, accesses to each sub-bank can be processed in parallel.  The gather/scatter instruction takes a base address and a number of offset addresses as input. 
The sums of them form the addresses of the data in the sub-banks to be gathered or scattered. 
If there is no bank conflict, {\it i.e. } no two or more addresses fall into the same bank, the throughput is one gather or scatter per cycle. If there is a bank conflict, such accesses needs to be serialized to different cycles, and the throughput is reduced accordingly. 

An example is illustrated in Figure~\ref{fig:overall_arch} to complete a length-16 element-wise product between a sparse weight vector w[] and a dense activation vector a[] (see Figure~\ref{fig:overall_arch}(b)) with one GS memory access and one length-4 SIMD multiply accumulation (see Figure~\ref{fig:overall_arch}(a)). Four non-zero weights in w[] and their indices idx[] are compactly stored in DRAM and prefetched through cache hierarchy to the SIMD processor. Those indices are used as the offsets to address 4 activation data in 4 different sub-banks, where a[] is sequentially stored across 4 sub-banks such that a[$i$] is stored in the ($i$ mod 4)th sub-bank. As long as the 4 entries in idx[] modulo 4 are unique (e.g., idx[]=\{4,7,13,14\}), the a[] {that} the indices pointed to can be loaded in one memory access. Then the processor can perform a length-4 SIMD product between (w[4],w[7],w[13],w[14]) and (a[4],a[7],a[13],a[14]).

The on-chip TCM and gather/scatter engine are widely used in many classical applications such as look-up table  (LUT)~\cite{lucente1993interactive}, Haar-Cascade~\cite{soo2014object}, and image warping~\cite{wolberg1990digital}. The efficient random data accesses of the gather/scatter engine on TCM make it also suitable for randomly referencing the activations of the deep learning models. 
\section{Gather-scatter Sparse Patterns}
\label{sec:GS_pattern}

\begin{figure*}[tbp]
\centering
  \centering
  \begin{center}
  \begin{tabular}[t]{@{} c c c@{} }
  \includegraphics[width=0.315\linewidth]{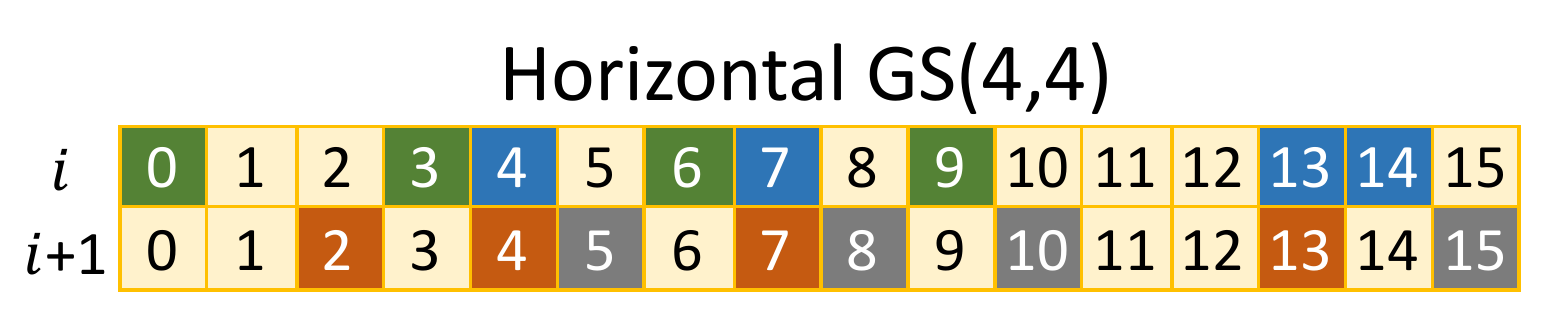}
  &
  \includegraphics[width=0.315\linewidth]{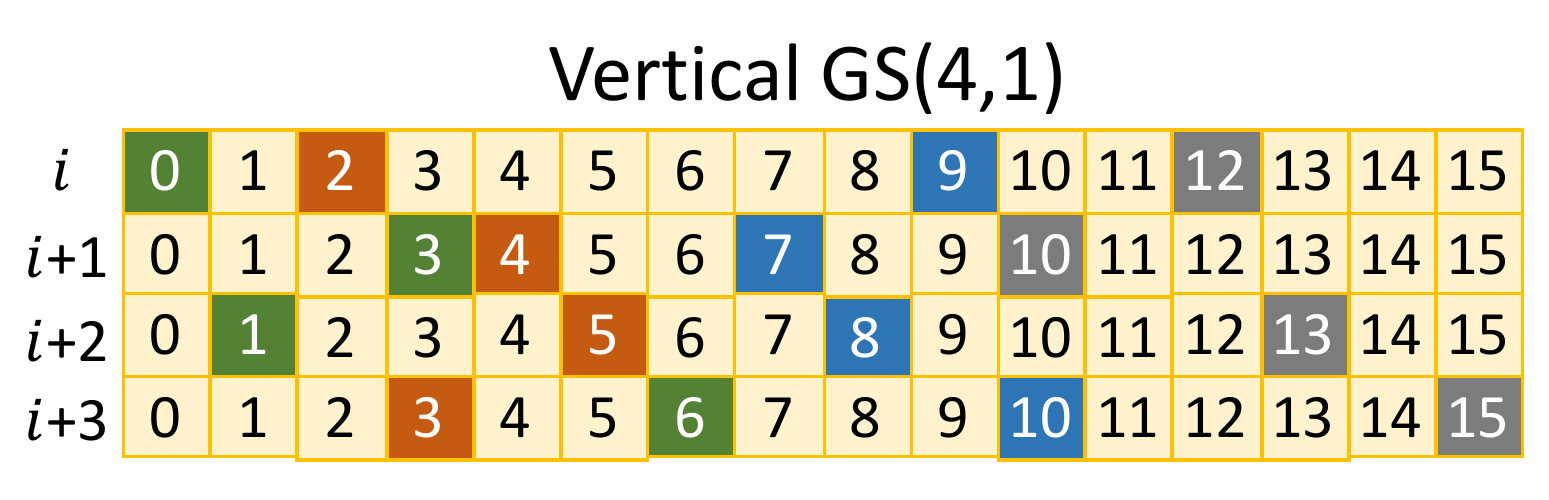}
  &
  \includegraphics[width=0.315\linewidth]{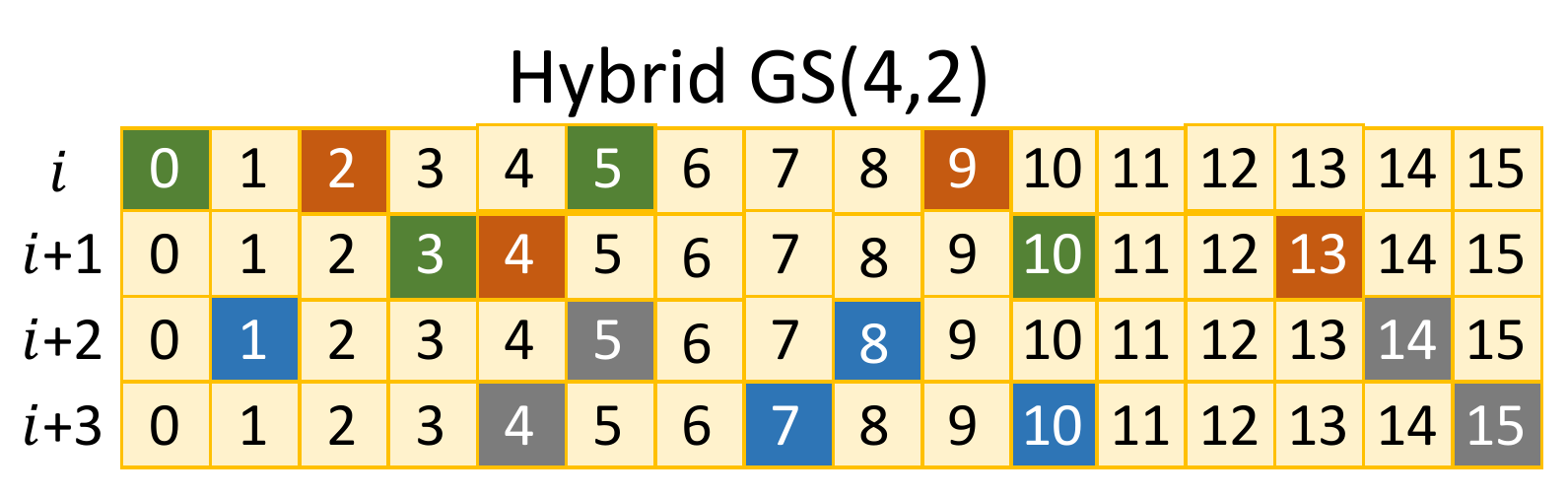}
  \\
  \small{(a)} & \small{(c)} & \small{(e)}\\
   \includegraphics[width=0.3\linewidth]{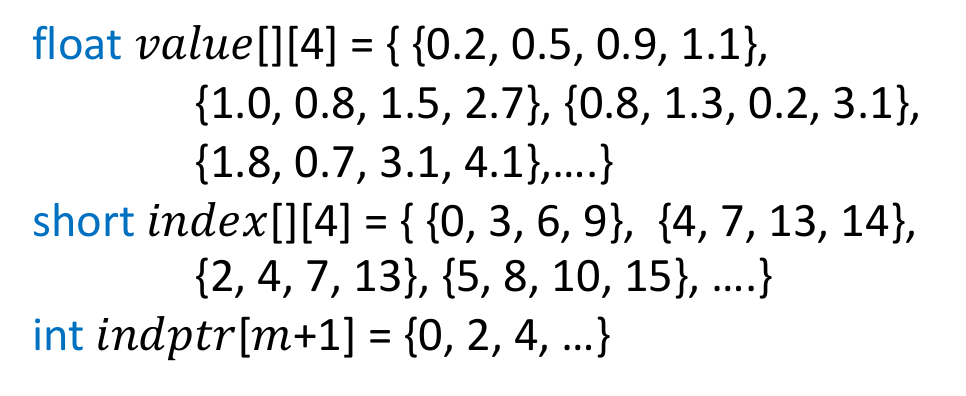}

  &
  \includegraphics[width=0.3\linewidth]{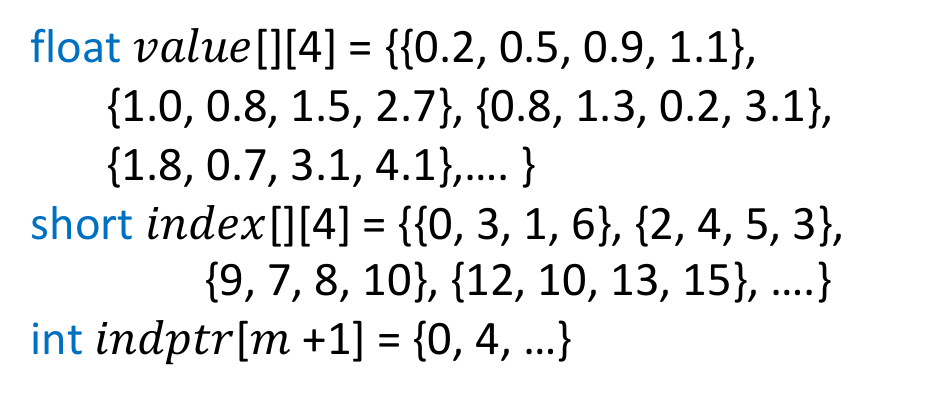}
  &
  \ifmaximal
  \includegraphics[width=0.315\linewidth]{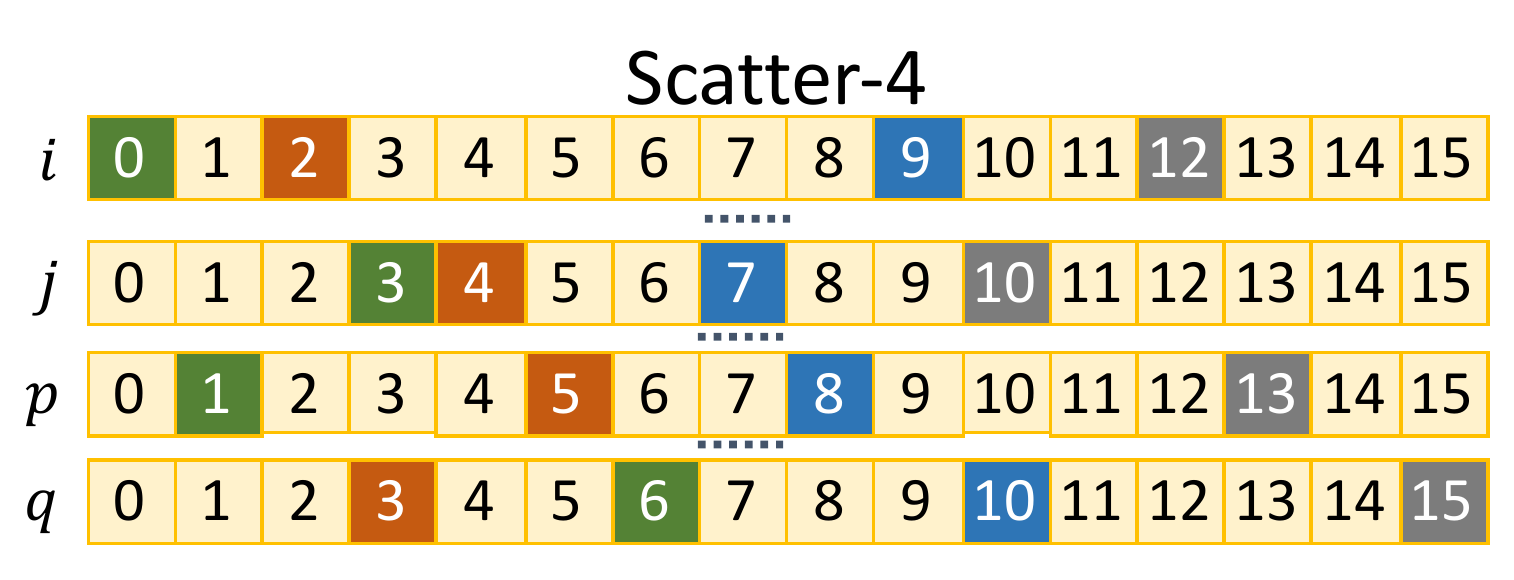}
  \fi
  \\
  \small{(b)} & \small{(d)} & \ifmaximal \small{(f)} \fi\\
  \end{tabular}
  \end{center}
  \caption{\small Examples of gather-scatter sparse patterns for the transposed weight matrices, with the number of sub-banks  being four. (a) Sparse weight matrix with GS horizontal patterns. (b) Compact representations of (a).  (c) Sparse weight matrix with GS vertical patterns. (d) Compact representations of (c). (e) Sparse weight matrix with GS hybrid patterns. \ifmaximal (f) Sparse weight matrix with GS scatter patterns. \fi }
  \label{fig:sparse_formats}
  
\end{figure*}

Directly applying gather-scatter engines to the irregular pattern (best theoretical sparsity pattern) with canonical sparse formats (e.g., compressed sparse row (CSR) or coordinate list (COO))  may not effectively take advantage of the TCMs and gather/scatter engines, as unconstrained irregularity may introduce  bank conflicts and reduce the data transfer throughput. As an example, in GNMT model~\cite{wu2016google} at 90\% irregular sparsity with activations saved in a 16-bank TCM, when CSR format is used in the ascending order of the indices, 2.8$\times$ accesses are needed over the perfectly balanced pattern. Even if the indices in a row are reordered to maximally reduce the bank conflicts, an extra 54\% accesses are needed. Thus, it is necessary to introduce a new set of sparse patterns to accommodate the characteristics of gather-scatter engines, which are described below.



\subsubsection{GS horizontal}
For the ease of illustration, assuming a dense vector of activation $a$[] is sequentially stored in a TCM with 4 sub-banks. $a[i]$ is stored in the $(i\bmod4)$-th sub-bank (see Figure~\ref{fig:overall_arch}). The computation of a neuron is the dot product of two vectors, activation $a$[] and weights $w$[], calculated by $\sum_i a[i]w[i]$. Note that $w$[] is a sparse vector. If we partition $w$[] into 4 groups based on its index $i\bmod4$, the gather-scatter engine and SIMD instruction can load and calculate $a[i]w[i]$ in parallel, as long as those $(i\bmod 4)$ are all different numbers (belonging to different sub-banks). Therefore, a maximum of 4 loads and multiplications can be completed with one memory access. The total number of accesses to compute $\sum_i a[i]w[i]$ is therefore the maximum number of non-zero weights among those 4 groups. For a fixed sparsity, the number of accesses is then minimized if all 4 groups have the same number of non-zero weights. Based on this observation, we propose a gather-scatter (GS) horizontal pattern, where in each row of a weight matrix, the number of non-zero weights in each group partitioned by their column indices modulo the number of sub-banks is the same. 

Figure~\ref{fig:sparse_formats}(a) shows two rows of an exemplar weight matrix satisfying the GS horizontal pattern. Elements colored as light yellow are zeros and others are non-zero weights. Activations corresponding to weights of the same color \ifmaximal (green, blue, orange and grey) \fi can be loaded and multiplied with the weights in parallel since the column indices modulo 4 of the same color are all different. For example in the $i$th row, four weights in blue boxes have column indices $\{4,7,13,14\} \equiv \{0,3,1,2\} \bmod 4$. 

\subsubsection{GS vertical}
A GS vertical pattern can be similarly defined (see Figure~\ref{fig:sparse_formats}(c)). One group of 4 elements of the same color are equally distributed across 4 consecutive rows and their column indices modulo 4 are all different. The number of elements in each row is therefore equal to the number of groups. For example, the four weights in the green boxes are in the column indices $\{0,3,1,6\} \equiv \{0,3,1,2\} \bmod 4$, and the matching activations can be loaded and multiplied in parallel. Note that those 4 multiplications correspond to 4 rows ({\it i.e.}, 4 neurons) so that they are partial sums of 4 independent dot products and are not reduced together. The reduction happens when the results of the next group of 4 multiplications ({\it e.g.}, values in orange boxes) are element-wise accumulated with the first partial sums.

\subsubsection{GS hybrid}
Note that all elements in one gather come from the same row in the GS horizontal pattern, and come from different rows in the GS vertical pattern.
They can be generalized to a GS hybrid pattern where the number of elements gathered per row is in-between and their column indices modulo the number of sub-banks are all different. 
In Figure~\ref{fig:sparse_formats}(e), the four elements gathered in one memory access are divided to two consecutive rows, with two elements in each row being gathered.

\ifmaximal
\subsubsection{GS scatter}
Another generalization is to relax the constraints on row consecutiveness. Figure~\ref{fig:sparse_formats}(f) shows a GS vertical pattern where the four rows can be arbitrary, and we call it GS scatter pattern. It further relaxes the constraints on GS vertical patterns to multiply the activations with four non-consecutive rows of weights and scatter the results.
\fi

The proposed GS patterns are formally defined as follows.

\begin{definition}\label{def:2d-gs}{\em
Let the weights of a layer be a 2-dimensional matrix $\mathbf{W} \in \mathbb{R}^{m\times n}$. Let the number of sub-banks of a gather-scatter engine be $B$. Let $k$ be a divisor of $B$. For simplicity, assume $m$ is divisible by $\frac{B}{k}$.
The $i$th row is denoted by $\mathbf{W}[i]$. 
All indices are zero-based numbering. 

Let the total number of non-zero elements in the sub-matrix formed by the consecutive rows from $\mathbf{W}[i]$ to $\mathbf{W}[i+\frac{B}{k}-1]$ be $N$. This sub-matrix is said to satisfy the {\em \textbf{ GS hybrid}} pattern with $k$ elements per row if the following two properties are satisfied,
\begin{itemize}
    \item The number of non-zero elements in each row are the same, i.e.,
    \begin{small}
    \begin{align*}
        | \left\{ (z,j) : W[z][j] \neq 0 \right\} | =  \frac{Nk}{B},
    \end{align*}
    \end{small}
    for all $z \in \{i,..., i+\frac{B}{k}-1\}$.
    \item The non-zero elements' column indices modulo $B$ are equally distributed in $\{0,1,\ldots, B-1\}$, {\it i.e.}, 
    \begin{small}
    \begin{align*}
        \sum_{z=i}^{i+\frac{B}{k}-1} | \left\{ (z,j) : W[z][j] \neq 0, j \equiv b \bmod B \right\} | = \frac{N}{B},
    \end{align*}
    \end{small}
    for all $b\in\{0,1,\ldots, B-1\}$.
\end{itemize}
}

\end{definition}

The weight matrix $\mathbf{W}$ is said to satisfy the GS hybrid pattern with $k$ elements per row if all $\frac{mk}{B}$ sub-matrix partitions formed by consecutive $\frac{B}{k}$ rows satisfy the GS hybrid pattern. We denote this hybrid pattern by $GS(B,k)$.
A weight matrix $\mathbf{W}$ is said to satisfy the {\em \textbf{ GS horizontal} } pattern if $\mathbf{W}$ satisfies $GS(B,B)$. A weight matrix $\mathbf{W}$ is said to satisfy the {\em \textbf{ GS vertical} } pattern if $\mathbf{W}$ satisfies $GS(B,1)$. 
\ifmaximal A weight matrix $\mathbf{W}$ is said to satisfy the {\em\textbf{GS scatter}} pattern if there exists a matrix $\mathbf{\widehat W}$ that is formed by permuting rows of $\mathbf{W}$, such that $\mathbf{\widehat W}$ satisfies $GS(B,k)$. It will be denoted by $GS_{\textrm{scatter}}(B,k)$. \fi

As a comparison, the block structured sparsity can be denoted in a similar manner. $Block(B,k)$ indicates that $B$ consecutive elements are non-zero, with $k$ elements along the row dimension, and $\frac{B}{k}$ elements along the column dimension. {\em \textbf{Block horizontal}} and {\em \textbf{Block vertical}} patterns can be represented as $Block(B, B)$ and $Block(B, 1)$ respectively.

\begin{definition}\label{def:3d-4d-gs}{\em
Let the weights of a layer be a 4-dimensional matrix $\mathbf{W} \in \mathbb{R}^{O\times h\times w \times I}$. The projection $f:\mathbb{R}^{O\times h\times w \times I}\rightarrow \mathbb{R}^{O\times (hwI)}$ is defined such that the last three dimensions are flattened to a 1-D vector of size $hwI$, where the most inner scanning order is in the $I$ dimension. Then $\mathbf{W}$ is said to satisfy GS horizontal, vertical, \ifmaximal hybrid, or scatter \fi pattern if $f(\mathbf{W})$ satisfy GS horizontal, vertical, \ifmaximal hybrid, or scatter \fi pattern, respectively.

Let the weights of a layer be a 3-dimensional matrix $\mathbf{W} \in \mathbb{R}^{O\times L \times I}$. The projection $f:\mathbb{R}^{O\times L \times I}\rightarrow \mathbb{R}^{O\times (LI)}$ is defined such that the last two dimensions are flattened to a 1-D vector of size $LI$, where the most inner scanning order is in the $I$ dimension. Then $\mathbf{W}$ is said to satisfy GS horizontal, vertical, \ifmaximal hybrid, or scatter \fi pattern if $f(\mathbf{W})$ satisfies GS horizontal, vertical, \ifmaximal hybrid, or scatter \fi pattern, respectively.
}
\end{definition}

\begin{figure}[tbp]
\centering
  \centering
  \begin{center}
  \begin{tabular}[t]{ @{}c c@{}  }
  \includegraphics[width=0.35\linewidth]{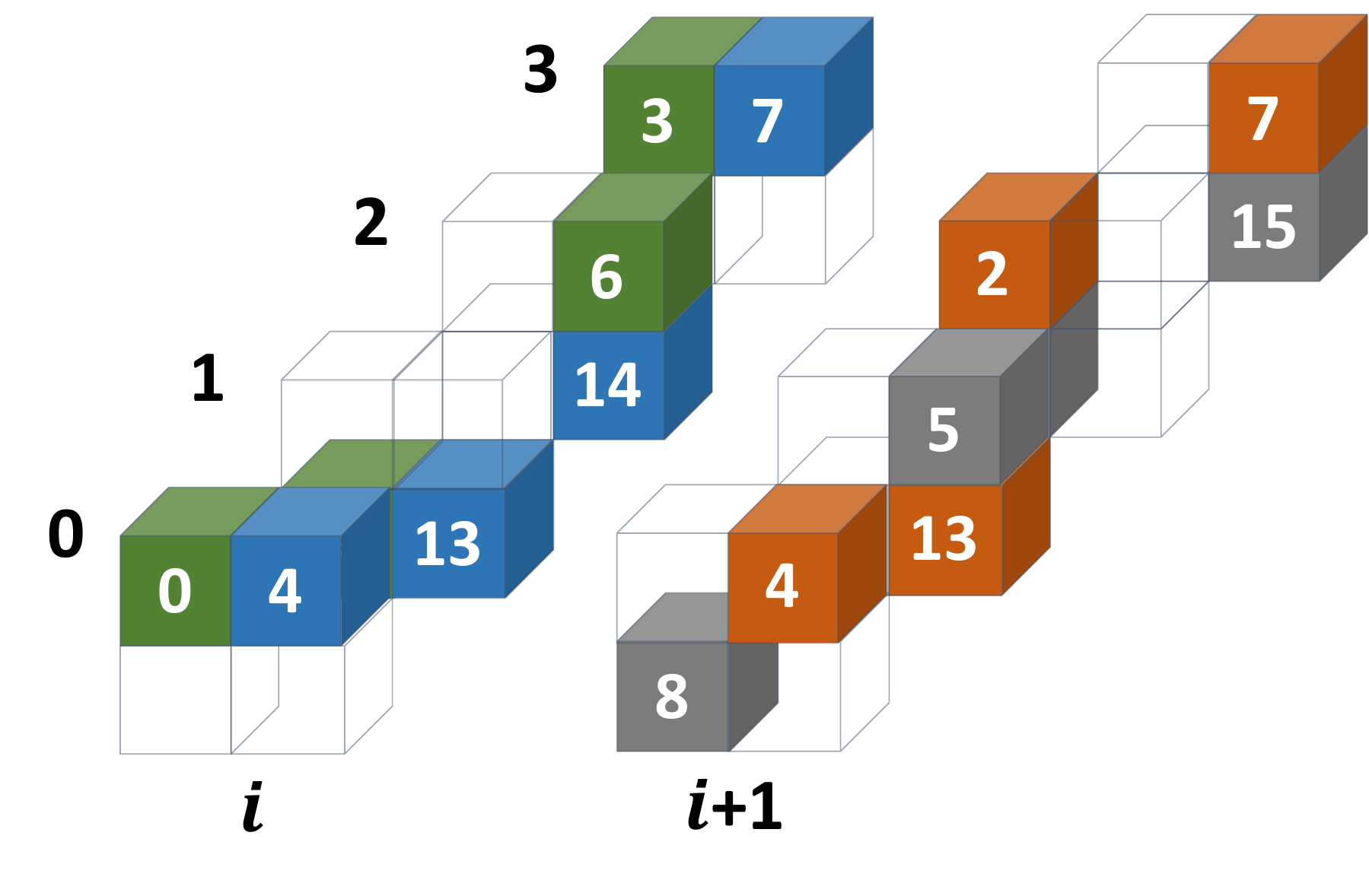}
 &
  \includegraphics[width=0.59\linewidth]{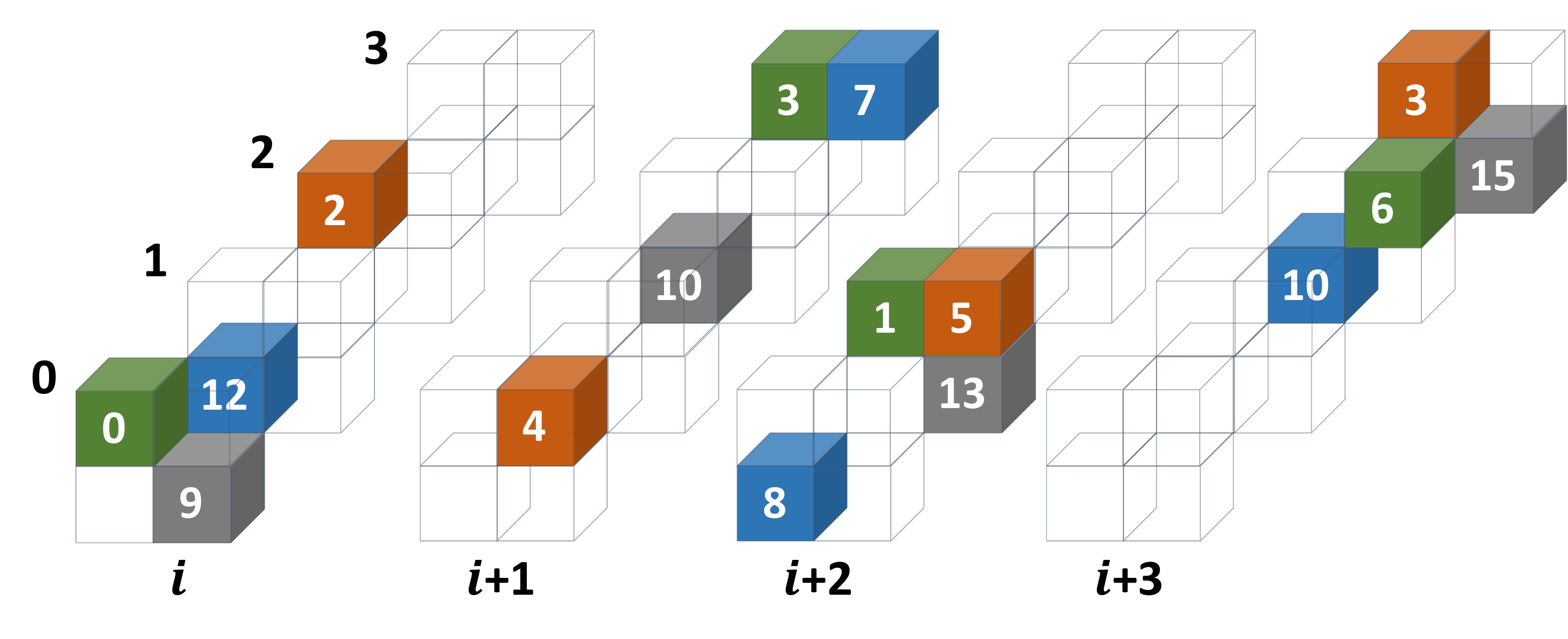}
  \\
  \small{(a)} & \small{(b)}
  \end{tabular}
  \end{center}
  \caption{\small Examples of bank balanced sparse format for 2x2 convolutions with four channels. The number of sub-banks is set to four. (a) Uncompressed format for the horizontal pattern. (b) Uncompressed format for the vertical pattern.}
  \label{fig:sparse_conv_patterns}
  
 \end{figure}

Definition~\ref{def:2d-gs} defines the GS patterns that complete spMV and spMM using gather-scatter 
\ifmaximal 
engines.
The applicable DNN models have 2-D weight matrices, such as MLP, LSTM, GRU, etc. 
\fi
Definition~\ref{def:3d-4d-gs} defines the GS patterns that complete sparse 1-D and 2-D convolutions. 
\ifmaximal
The applicable DNN models have 1-D and 2-D convolutional layers, where the first $O$ dimension indicates the output channels, the next 1 or 2 dimensions indicate the 1-D or 2-D convolutional kernel size, and the last $I$ dimension indicates the input channels.
\fi
Figure~\ref{fig:sparse_conv_patterns} shows an example of 2-D convolution, where a 3-D filter, formed by four $2 \times 2$ convolutional kernels, is projected onto a row vector in Figure~\ref{fig:sparse_formats}. For example, the two filters in Figure~\ref{fig:sparse_conv_patterns}(a) satisfy the GS horizontal pattern and are mapped to the two consecutive rows in Figure~\ref{fig:sparse_formats}(a); the four filters in Figure~\ref{fig:sparse_conv_patterns}(b) satisfy the GS vertical pattern and are mapped to the four consecutive rows in Figure~\ref{fig:sparse_formats}(c).

\section{Sparse Computation Using GS patterns} \label{sec:sparse_computation}
 Compact representations of GS patterns can reduce the memory cost of storing the weight values in the cache while containing the addresses to locate the corresponding activations in the TCMs via GS engines.  We propose a sparse format similar to the block compressed row (BSR) format \ifmaximal that encodes the information of the TCM banks \fi for efficient model inference. 
 We use row-major order to describe the format, 
 even though the format is major agnostic.
 \ifmaximal
 We first present the compact formats and the sparse computation algorithms for spMV applicable to models consisting of MLPs, LSTMs, GRUs etc., and then generalize them to sparse 1-D and 2-D convolutions, which are widely used in vision and acoustic models.
\fi

\subsection{Sparse matrix   dense   vector   multiplication}
The format is composed of three arrays: $value$, $index$, and $indptr$ arrays.
The $value$ and $indptr$ arrays are the same as the BSR format. 
The $value$ array is a two-dimensional array, where the first dimension is the number of non-zero groups, and the second dimension is the number of values in a group\ifmaximal ({\it i.e. } a block of consecutive values)\fi. The number in the second dimension matches the number of sub-banks in the TCM. 
The value in each entry of the $indptr$ array is the number of non-zero groups from the first row up to the row of the index of the entry. 
\ifmaximal
In BSR format, the $index$ array is a one-dimensional array containing the number of non-zero groups, with the value of each entry being the column index of the corresponding block in the $value$ array. 
\fi
Different from the BSR format, 
the $index$ array is a two-dimensional array\ifmaximal similar to the $value$ array\fi, where the first dimension is the number of non-zero groups, and the second dimension is the number of values in a group. Its values contain the column numbers for the corresponding entries in the $value$ array. 
The values are ordered such that the column indices in a group modulo the number of entries of the group  are unique. 
Figure~\ref{fig:sparse_formats} (b) and (d) show two examples of the proposed format, which represent the GS horizontal and vertical patterns shown in Figure~\ref{fig:sparse_formats} (a) and (c). 
In both examples, the TCM contains four sub-banks. Four individual values with no bank conflict can be gathered or scattered from or to the TCM.


\begin{algorithm}
\caption{spMV in horizontal format.}
\label{alg:2-d-horizontal}
\begin{small}
    \KwIn{act, value, index, indptr}
    \KwOut{result}
    \For {(i = 0; i $<$ m; i++)} {
        intx4 res = 0\;
        \For {(idx = indptr[i]; idx $<$ indptr[i+1]; idx++)} {
            shortx4 weight = value[idx]\;
            shortx4 ofst = index[idx]\;
            shortx4 activation = GATHER(\&act, ofst)\;
            res += activation  $\odot$ weight\;
        }
        result[i] = REDUCTION(res)\;
    }
\end{small}
\end{algorithm}

We describe the pseudo-code of the spMV algorithm for the GS horizontal pattern in Algorithm~\ref{alg:2-d-horizontal}. We use 4 TCM sub-banks as an example and it is straightforward to generalize to a larger value.
The inputs to the pseudo-code are the activation vector $act$, the $value$, $index$, and $indptr$ arrays of the sparse weight matrix. The output is the resulting activation vector $result$ of length $m$ (the number of rows in the weight matrix). In Line~1, the kernel iterates over all output entries. The inner loop (Line~3-8) iterates over all groups in one row of the weight matrix, with one load of the weight values (Line~4), one load of the column indices (Line~5), and one gather of the activation entries (Line~6). Then the activation and weight entries are element-wise multiply-accumulated in the $res$ variable (Line 7). Lastly, the four entries in the $res$ variable needs to be reduced to one scalar value and saved to the $result$ vector (Line~9). 
As an optimization, the $value$ and the $index$ arrays can be merged to one joined array, which has better cache locality characteristics.
Comparing with the scalar implementation, the number of iterations in the outer loop is unchanged ($m$), while the number of iterations in the inner loop is reduced by 4 times, where 4 equals the number of TCM sub-banks.

\ifmaximal

\begin{algorithm}
\caption{spMV in vertical format.}
\label{alg:vertical}
\begin{small}
    \KwIn{act, value, index, indptr}
    \KwOut{result}
    \For {(i = 0; i $<$ N / 4; i++)} {
        intx4 res = 0\;
        \For {(idx = indptr[i]; idx $<$ indptr[i+1]; idx++)} {
            shortx4 weight = value[idx]\;
            shortx4 ofst = index[idx]\;
            shortx4 activation = GATHER(\&act, ofst)\;
            res += activation $\odot$ weight\;
        }
        result[i * 4] = res\;
    }
\end{small}
\end{algorithm}

In the GS vertical pattern, the four elements in a bundle come from four consecutive rows in the weight matrix, as shown in Figure~3~(c-d). The pseudo-code of the corresponding spMV computation is shown in Algorithm~\ref{alg:vertical}. One difference from the horizontal sparse format is that the number of iterations in the output loop is reduced to a quarter ($N/4$), while the number of iterations in the inner loop becomes the maximum of the number of non-zero entries in the four rows. In addition, the vector value of $res$ in Line~9 already contains the results for the four rows so no reduction is needed for vertical sparse format. When using this format, all rows in a bundle contain the same the number of non-zero entries.

\fi

The spMV algorithm for GS \ifmaximal hybrid, and scatter \fi patterns can be designed similar to Algorithm~\ref{alg:2-d-horizontal}. \ifmaximal In addition, the GS scatter pattern requires a fourth array to indicate the entries of the outputs. Since in deep learning, the number of non-zero elements is much larger than the number of rows in the weight matrices, the extra storage overhead is  negligible. \fi


\subsection{Sparse Convolution}
Similar to spMV computation, CNN models can also take advantages of the gather-scatter engine and GS patterns via sparse convolutions. 
We assume the convolution filters to be in the $O h w I$ layout as in Definition~\ref{def:3d-4d-gs}.
It matches the kernel layout of $NHWC$, where $N$ is the inputs in a batch; $H$ and $W$ are the height and width of the activation; and $C$ is the input channels.
\ifmaximal
Note that the $OIhw$ layout can also be specified, though the calculation of the entries in the $index$ array is different. 
\fi

Figure~\ref{fig:sparse_conv_patterns} (a) shows two uncompressed 2x2 sparse convolution filters with 4 input channels each. The non-zero entries are the same as the one presented in Figure~\ref{fig:sparse_formats} (a). Since the input channel $I$ is the innermost {dimension}, each input channel is mapped to a separate TCM sub-bank, and can be gathered in parallel. In the convolution computation, the sparse filter moves along the $HW$ dimensions of the activations (feature maps) and performs dot product with matching elements. Since the matching elements in the kernel for the second row of the filter ({\it i.e.} $h = 1$) are not immediately following the first row of the filter, the corresponding entries in the $index$ array are not the same as the one described in Figure~\ref{fig:sparse_conv_patterns} (b). Instead, the indices for the filter entries in the second row need to be offset by $(W-w)C$. 
As an example, the indices for the first group in Figure~\ref{fig:sparse_conv_patterns} (a) is $\{0, 3, 6, WC + 1\}$. Thus, sparse convolutions specified using this format is kernel shape aware. 

\section{Pruning Methodology}
The sparse patterns and formats 
described in Section~\ref{sec:GS_pattern} and Section~\ref{sec:sparse_computation} 
present specifications to avoid bank conflict. This section presents {the} pruning techniques to satisfy the specifications during training.


\begin{algorithm}
\caption{Horizontal sparse pattern selection.}
\label{alg:sparse_pattern}
\begin{small}
    \KwIn{weight, sparsity, B}
    \KwOut{pattern}
    buckets = allocate B lists\;
    threshold = percentile(abs(weight), sparsity)\;
    \ForEach {row} {
        empty buckets\;
        \ForEach {col} {
            bnum = col mod B\;
            buckets[bnum].append((weight[row][col], col))\;
        }
        \For {(i in 0..B)} {
            sort(buckets[i][j] on abs(buckets[i][j][0]))\;
        }
        num\_items = sum(abs(weight[row]) $>$ threshold)\;        
        \For {(; num\_items $>$ 0; num\_items -= B)} {
            \For {(i in 0..X)} {
                top\_entry = buckets[i][0].pop()\;
                pattern[row, top\_entry[1]] = 1\;
            }
        }
    }
\end{small}
\end{algorithm}

Algorithm~\ref{alg:sparse_pattern} describes the pseudo-code for the horizontal pattern {selection}.
We allocate $B$ buckets where $B$ is the number of sub-banks (Line 1). Then we calculate the threshold value of the weight based on the sparsity level as if the pattern is irregular (Line 2). 
Since the pattern is horizontal, each row of the weight is considered independently.
In every {\em row}, for each {\em col} of the weight matrix, we append the tuple of weight value and column index $(v, c)$ to the bucket number $c \bmod B$ (Line 5-8). Then we sort the entries in the buckets in descending order of the {absolute values of the} weight {$abs(v)$} (Line 9-11). After that, we repeatedly pick the first entry in each bucket to form a group until the desired number of non-zero values have been collected (Line 13-18). This forms the horizontal pattern for one row. This process is repeated for all rows in the weight matrix. 

The vertical and hybrid patterns are similar. The group structure is still allocated in every row. When we pick entries to form a group, we greedily search all rows in a group and pick the first bucket entry with the maximum absolute weight value in the available bucket pool. We repeat the process until all buckets in a group are filled. 

\ifmaximal
The scatter pattern is similar to the vertical or hybrid patterns. Instead of forming a group from the data in consecutive rows, we first sort all rows in descending order based on the number of entries above the threshold calculated using the irregular pattern. Then we group entries from the neighboring sorted rows using the same method as the vertical or hybrid patterns. In this case, the number of non-zero entries in neighboring sorted rows are similar so the imbalance in the groups is minimized. 
\fi

\section{Experimental Results}

\begin{figure*}[t]
{\small 
\centering
\begin{tabular}[t]{ c c c  }
  \includegraphics[width=0.31\linewidth]{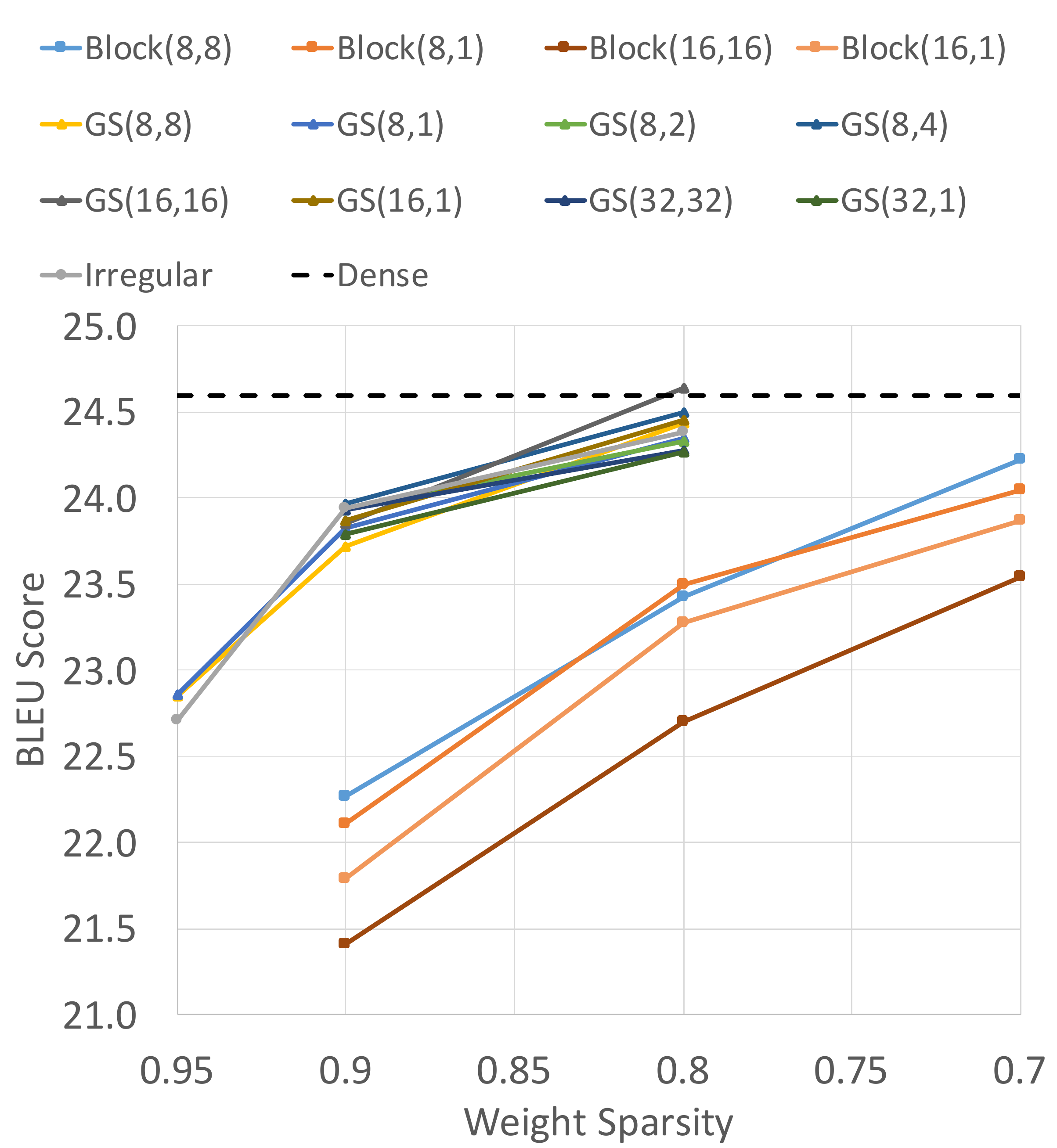} & 
  \includegraphics[width=0.31\linewidth]{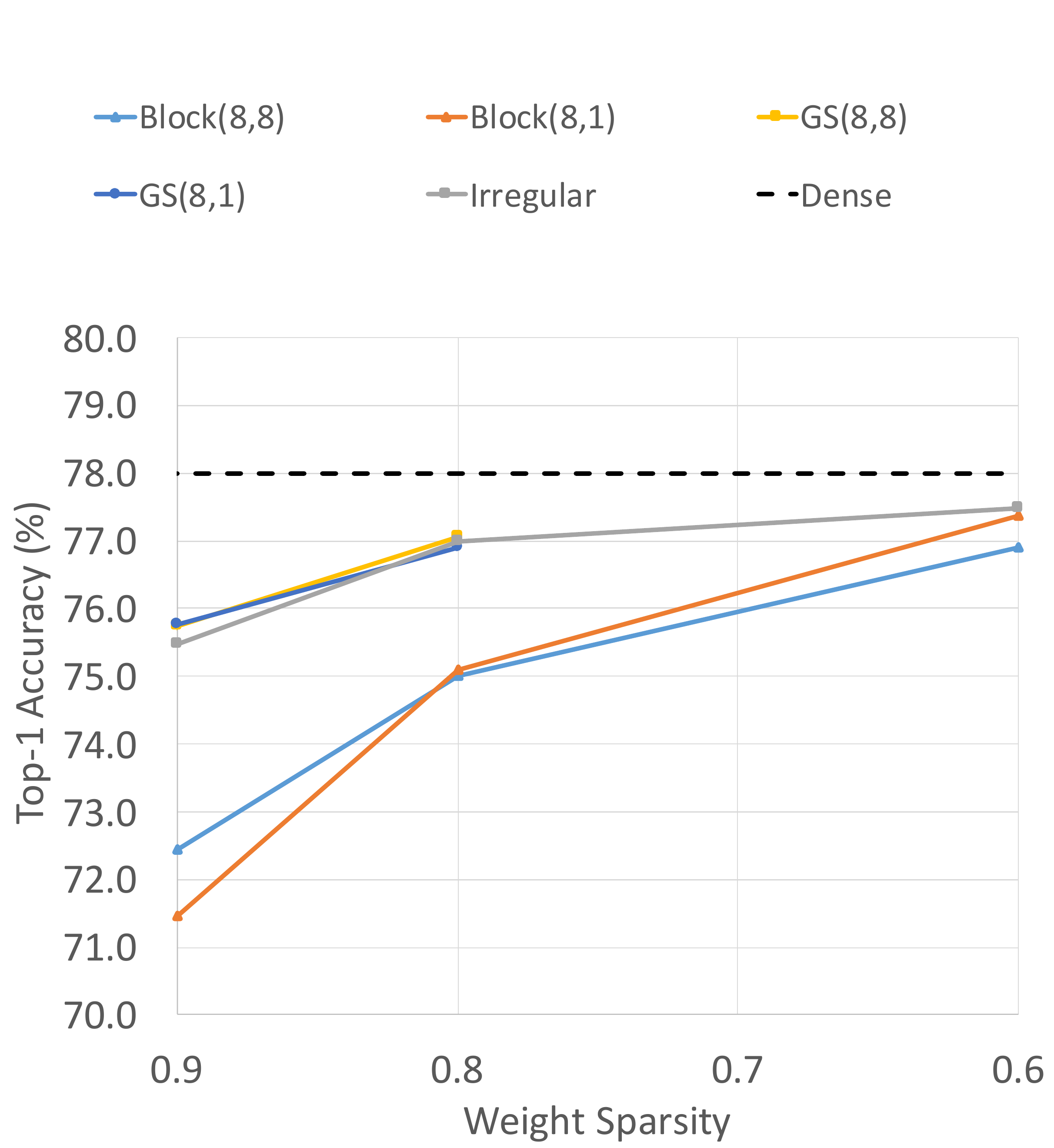} &
  \includegraphics[width=0.31\linewidth]{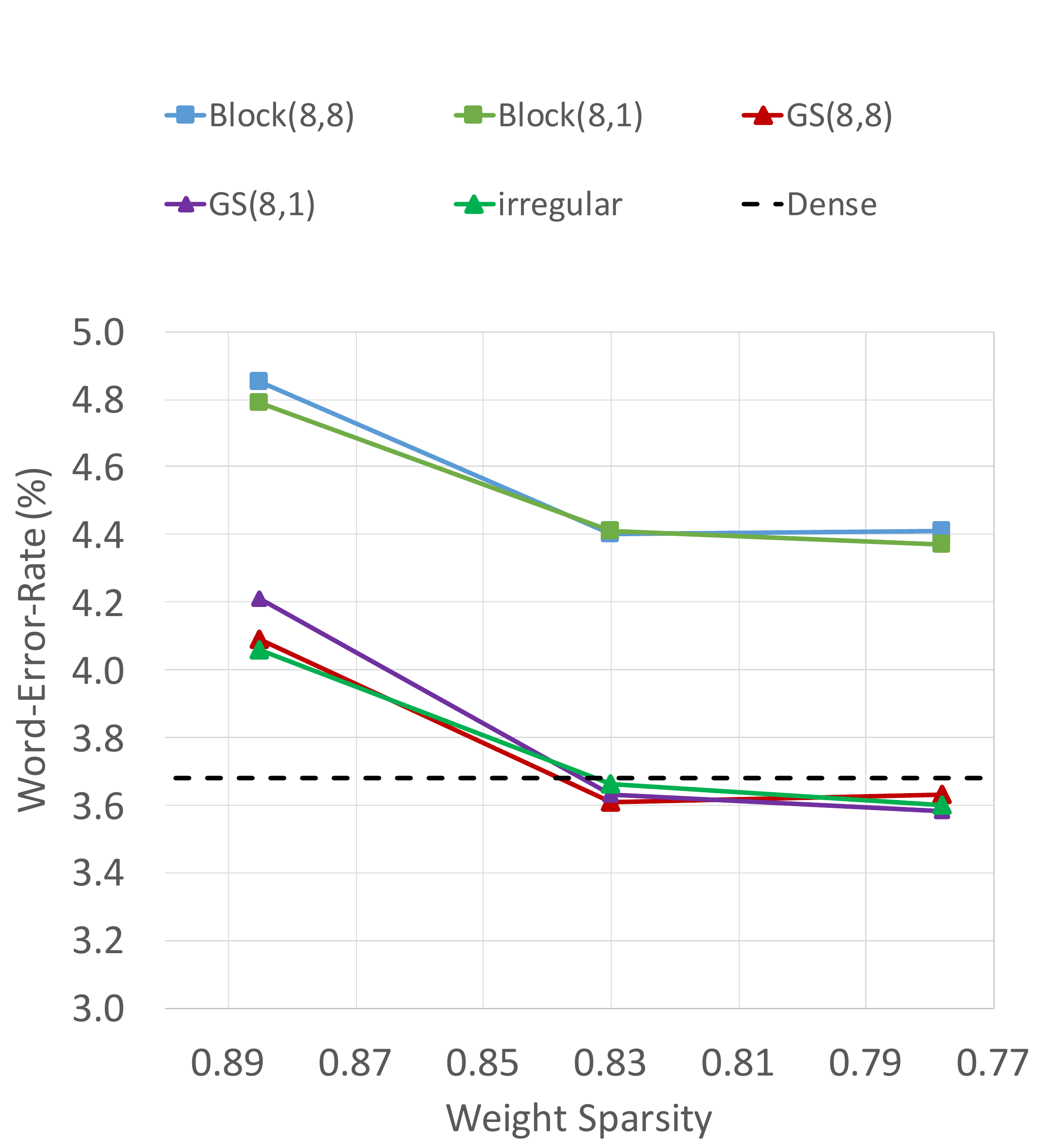} \\
  
  \small{(a) GNMT} & \small{(b) ResNet50} & \small{(c) Jasper} 
  
 \end{tabular} 
  \caption{Quality comparison of models with irregular, GS, and block sparse patterns. (a) shows the BLEU scores (y-axis) and the weight sparsity (x-axis) for GNMT models. (b) shows the top-1 accuracy (y-axis) and weight sparsity (x-axis) for ResNet50 models. (c) shows the word-error-rate (y-axis) and weight sparsity (x-axis) for Jasper models.}
  \label{fig:gs_results}
  }
\end{figure*}

\subsection{Model accuracy}
\label{sec:exp_model_quality}
We demonstrate the benefits of the proposed gather-scatter sparse patterns on 3  different DNN architectures for different deep learning tasks.
Most of the hyper-parameters of our experiments follow the public repository of NVIDIA at~\cite{NVIDIA}.
Detailed experimental setup and numerical results are provided in the supplementary material.


\subsubsection{GNMT~\cite{wu2016gnmt} on machine translation}


Figure~\ref{fig:gs_results}(a) demonstrates the comparison of irregular, GS and block sparse patterns. All layers have the same sparsity when pruned. The x-axis is the weight sparsity and
the y-axis is the BLEU~\cite{papineni2002bleu} scores. Hence, results in the upper left of the figure are desired. 


We observe that a) GS sparse patterns can maintain the BLEU score comparing to the dense GNMT model at 80\% sparsity level. b) BLEU scores with GS and irregular patterns (no constraints) are similar. c) Comparing GS and block patterns at the same sparsity level, GS sparse patterns are 1.1 point and 1.5 points higher than block sparse patterns, respectively. d) Comparing GS and block patterns at the same BLEU scores, GS patterns can compress the GNMT models 2-3 times more than the block patterns.

\subsubsection{ResNet-50~\cite{he2016deep} on 2-D image recognition}

Figure~\ref{fig:gs_results}(b) demonstrates the comparison of the sparse ResNet-50 models. All layers have the same sparsity when pruned. The y-axis is the top-1 accuracy,
and results in the upper left of the figure are desired. 
We observe that a) GS patterns maintain {the} accuracy at 80\% sparsity and they are similar to the irregular pattern where no constraints are imposed. b) Comparing the GS and block patterns at the same sparsity level, {the} GS patterns  offer 2\% and 4\% accuracy improvement. This improvement is projected to be even larger for more sparse models. c) Comparing the GS and block patterns at the same accuracy, GS patterns can offer more than twice the compression ratio than the block patterns.


\subsubsection{Jasper~\cite{Li2019JasperAE} on speech recognition}

Figure~\ref{fig:gs_results}(c) demonstrates the comparison of word-error-rate (WER in percentage) of the sparse Jasper models. The x-axis is the overall weight sparsity. 
The y-axis is WER and results in the lower left of the figure are desired. 
We observe that a) GS and irregular patterns perform similar and they  both maintain the WER to the dense counterpart at 83\% of sparsity. b) Comparing GS and block patterns at the same sparsity level, GS offers 0.8\% less WER than the block patterns. c) At the same WER, the GS pattern offers twice compression ratio compared to the block pattern.



\subsection{Runtime performance}

In addition to the theoretical FLOP reduction, 
we also compare the runtime performance of the sparse kernels using the proposed patterns with the block sparse patterns and the dense kernels.
We set up the experiment system using GEM5~\cite{binkert2011gem5} simulator, which is a cycle accurate simulator widely used in computer architecture research.
The detailed experiment setup is described in the supplementary material. 

\begin{figure}[t]
{\small 
\centering
  \includegraphics[width=0.95\linewidth]{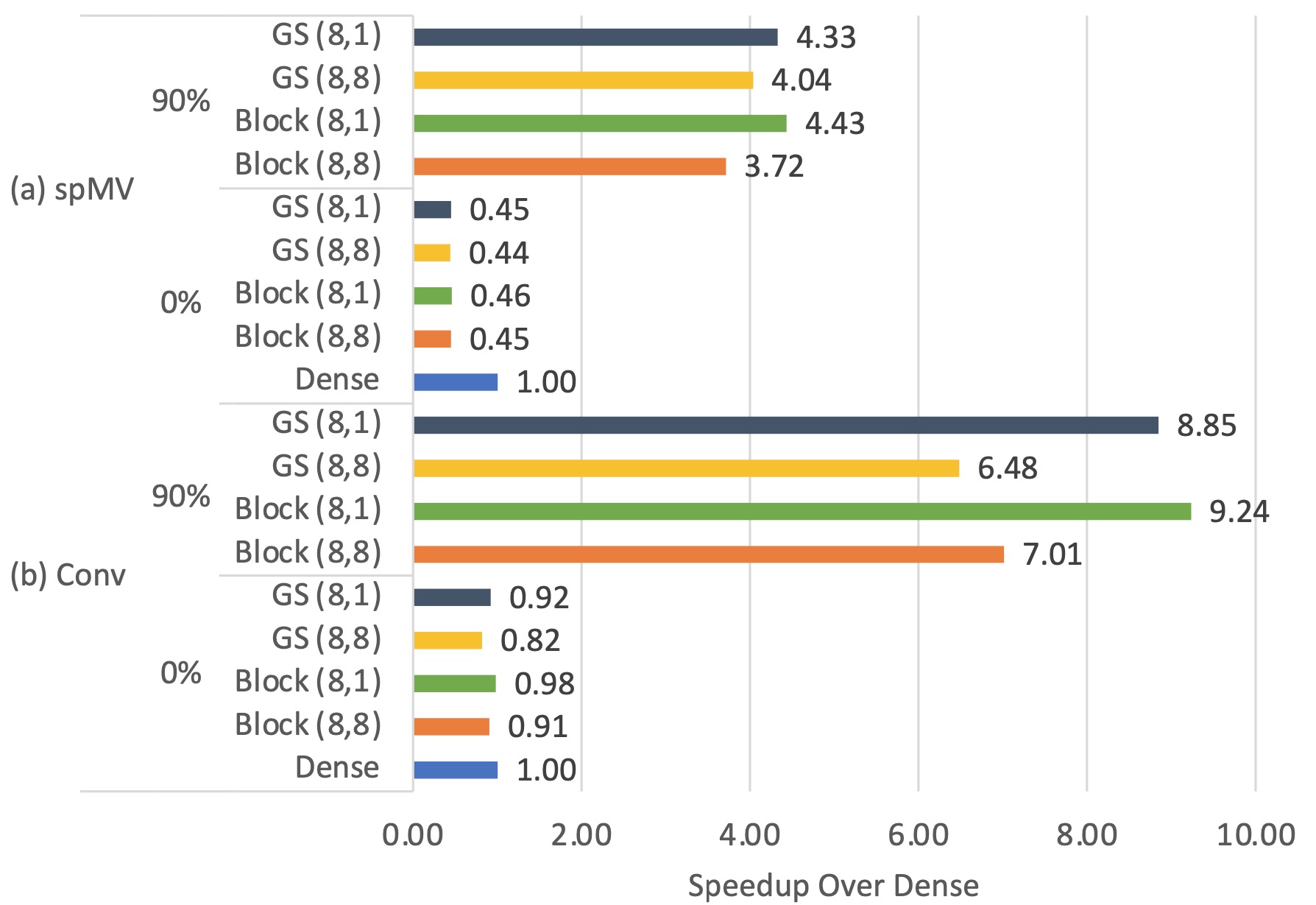}
  \caption{\small Kernel speedup of the block and gather/scatter patterns over the dense kernel at 0\% and 90\% sparsity levels. (a) spMV computation, and (b) sparse convolution.}
  \label{fig:kernel_runtime}
  }
\end{figure}

 We have spent considerable effort implementing and optimizing the kernels in the GS and block patterns. We compare the relative speedups over the dense kernel executing an $(1, 1024)\times(1024, 1024)$ spMV computation in Figure~\ref{fig:kernel_runtime} (a).
 The bar chart of the zero percent sparsity illustrates the efficiency of the sparse representation patterns. It is understandable that all sparse patterns are less efficient than the dense pattern, but the GS patterns are similarly as efficient as the block patterns. At 90\% sparsity, we use the real weight distribution in the decoder attention layer in the GNMT model. From it, we can see that the GS horizontal and vertical patterns can achieve speedups of $4.04\times$ and $4.33\times$, respectively. Its average speedup is $4.19\times$, slightly higher than the average speedup of the block sparsity pattern ($4.08\times$). 
 Similarly, Figure~\ref{fig:kernel_runtime} (b) compares the speedup on a convolution with {a} feature size of $8\times8$ and filter size of $3\times3$, with 128 input and output channels. 
 The average speedup of the GS and block patterns are $7.67\times$ and $8.13\times$, respectively. The performance of the GS patterns is only degraded by less than 5\%.
 \ifmaximal Higher speedup is achieved in sparse convolution than the spMV computation due to more data reuse.\fi
 In both experiments, the vertical patterns are more efficient than the horizontal patterns, which is because of its higher number of iterations in the inner loop.
 
 The experiments have shown that the kernels in the GS patterns deliver similar performance as the kernels in the block patterns. However, as shown in Figure~\ref{fig:gs_results}, higher sparsity levels can be achieved using the GS patterns at the same model quality. When the two factors are combined, the GS patterns can achieve $2-3\times$ speedup over the block patterns. This illustrates the effectiveness of the proposed method.
 
\section{Related Works}

Many network pruning methods have been proposed in order to reduce the number of non-zero values in a neural network~\cite{hassibi1993second,LeCun1990BrainDamage}. 
A magnitude-based iterative pruning method for {\em irregular} sparse patterns is proposed in~\cite{han2015learning}. $\ell_0$-based regularization and its $\ell_1$ relaxation methods are proposed in ~\cite{LearningL02017iclr} and~\cite{wen2016learning}. Other methods closely related to optimization theory are proposed to improve the sparsity on $\ell_0$ regularization in~\cite{zhang2018systematic,ren2019admm}. {\em Structured} pruning is more friendly to accelerators at the cost of model accuracy. For example, rows and columns in matrix-matrix multiplications can be removed by filter-wise pruning and shape-wise pruning~\cite{wen2016learning}.


In previous works, maintaining accuracy and hardware-friendly sparse patterns are not both achievable.
The sparsity patterns we propose in this paper can achieve the accuracy of irregular pattern while {being more} hardware friendly.

On the hardware side, fine-grained structured sparsity has gained a lot of attraction recently. 
Similarly, The NVIDIA A100 GPU~\cite{nvidia2020a100} achieves 50\% sparsity by choosing two out of every four elements. The density-adaptive regular-block pruning method is proposed in~\cite{ren2020darb} and has achieved higher sparsity levels on custom hardware. Bank balanced pruning method is proposed in~\cite{cao2019efficient}, focusing on spMV computation. 
Those works, however, are built on custom hardware accelerators. They explore sparsity among neighboring elements and are difficult to extend varying sparsity to an entire weight matrix row or an entire convolution filter. 
They also only focus on reducing the load imbalance along the reduction dimension ({\it i.e.} the horizontal pattern). In our work, we explore load imbalance in several different dimensions ({\it e.g.} horizontal, vertical, hybrid\ifmaximal, scatter\fi) across several weight matrix rows or convolution filters.
Our work is built on existing common modules in processors and DSPs, which is easier to be adopted. 
\section{Conclusion}

Sparse deep neural networks can avoid the redundant computation on zeros. But the random accesses and the load imbalance of the irregular sparsity have prevented it from being used in processors. 
In this paper, we efficiently use the existing gather/scatter engines in the processors. We present several novel sparse patterns and compact sparse formats specifically designed for the gather/scatter engines on random memory accesses. We also propose new pruning criteria to balance the loads among different TCM sub-banks. Our experiments on several models (GNMT, ResNet, Jasper) have shown that the resulting fine-grained sparsity patterns can achieve the quality close to irregular sparsity and the runtime efficiency on par with the block structured sparsity. The combined effect can result to $2-3\times$ runtime latency reduction over the block structured sparsity.




\clearpage

\section{Details on Experiment Setup}

{\small
\begin{table*}[th]
	\caption{Accuracy of the dense and sparse GNMT, ResNet50, and Jasper models of different sparse pattern: horizontal or vertical block pattern, horizontal or vertical gather-scatter (GS) pattern, and irregular pattern. GNMT reports BLEU scores, ResNet50 reports top-1 accuracy, and Jasper reports word error rate (WER).}
	
	\centering
		\begin{tabular}{c|c|c|c}
			\toprule
			 \begin{tabular}{c}
          Base \\ model 
        \end{tabular} &   Sparsity & Type & Score \\ \hline
			 \multirow{18}{4em} {GNMT} & 0\%  & Dense~\cite{wu2016gnmt} & 24.60 \\
			      & 70\% & Block(8,8)/(8,1)     & 24.22/24.05 \\
			      & 80\% & Block(8,8)/(8,1)     & 23.43/23.50 \\
			      & 90\% & Block(8,8)/(8,1)     & 22.27/22.11 \\
			      & 70\% & Block(16,16)/(16,1)     & 23.54/23.87 \\
			      & 80\% & Block(16,16)/(16,1)     & 22.70/23.28 \\
			      & 90\% & Block(16,16)/(16,1)     & 21.41/21.79 \\			      
			      & 80\% & Irregular & 24.38 \\
			      & 90\% & Irregular & 23.94 \\
			      & 95\% & Irregular & 22.71 \\
			      & 80\% & \textbf{GS(8,8)/GS(8,1)}  & \textbf{24.44/24.35} \\
			      & 80\% & \textbf{GS(8,2)/GS(8,4)}  & \textbf{24.33/24.50} \\
			      & 90\% & \textbf{GS(8,8)/GS(8,1)}  & \textbf{23.72/23.83} \\
			      & 90\% & \textbf{GS(8,2)/GS(8,4)}  & \textbf{23.93/23.97} \\
			      & 95\% & \textbf{GS(8,8)/GS(8,1)}  & \textbf{22.85/22.86} \\ 
			      & 80\% & \textbf{GS(16,16)/GS(16,1)}  & \textbf{24.64/24.45} \\
			      & 90\% & \textbf{GS(16,16)/GS(16,1)}  & \textbf{23.85/23.87} \\
			      & 80\% & \textbf{GS(32,32)/GS(32,1)}  & \textbf{24.28/24.27} \\
			      & 90\% & \textbf{GS(32,32)/GS(32,1)}  & \textbf{23.93/23.79} \\\hline
			\multirow{8}{4em} {ResNet50} & 0\%  & Dense & 78.02 \\
			                           & 60\% &  Block(8,8)/(8,1) & 76.91/76.98 \\
			                           & 80\% & Block(8,8)/(8,1)     & 75.02/75.10 \\
			                           & 90\% & Block(8,8)/(8,1)    & 68.28/71.47 \\ 
			                           & 80\% & irregular & 76.86 \\
                    			       & 90\% & irregular & 75.48 \\
                    			       & 80\% & \textbf{GS(8,8)/GS(8,1)}   & \textbf{77.06/76.91} \\
                    			       & 90\% & \textbf{GS(8,8)/GS(8,1)}  & \textbf{75.72/75.77} \\ \hline
            \multirow{9}{4em} {Jasper} & 0\%  & Dense & 3.68 \\
                                       & 77.8\% & Block(8,8) & 4.41 \\
                                       & 83\% &  Block(8,8)& 4.4 \\
                                       & 88.5\% &  Block(8,8) & 4.85 \\
                                       & 77.8\% & irregular & 3.60 \\
                                       & 83\% & irregular & 3.65 \\
                                       & 77.8\% & \textbf{GS(8,8)/GS(8,1)}  & \textbf{3.63/3.58} \\
                                       & 83\% & \textbf{GS(8,8)/GS(8,1)}  & \textbf{3.61/3.63} \\
                                       & 88.5\% & \textbf{GS(8,8)/GS(8,1)}  & \textbf{4.09/4.21} \\
                                       
			\bottomrule
		\end{tabular}
		
	\label{tab:accuracy-vs-sparsity}
\end{table*}
}

\subsection{Model Training Setup}
All models are trained using PyTorch~\cite{paszke2019pytorch} framework. If not explicitly mentioned, the hyper-parameters of the experiments follow the public repository of NVIDIA at~\cite{NVIDIA}.

\subsubsection{GNMT models on machine translation}

The Google neural machine translation (GNMT)~\cite{wu2016gnmt} model consists of four layers of LSTMs for both encoder and decoder. It also has attention layers and a classification layer. The sparsity is applied to all matrix multiplications inside the LSTMs, attention layers, and classifiers. The embeddings remain dense, as they are not weights and do not contribute to computing FLOPs.  

All sparse models are obtained by the prune-from-dense methodology with the ADMM~\cite{zhang2018systematic} regularization and then retrained. All layers are pruned to the same sparsity level. Each ADMM and retraining phase consists of 12 epochs using an ADAM optimizer and an initial learning rate of 0.00025. We train and evaluate on the WMT En $\rightarrow$ De dataset.

The first row in Table~\ref{tab:accuracy-vs-sparsity} lists the BLEU scores of the sparse GNMT models with different sparse patterns and sparsity levels. The model is one-shot pruned to 80\% sparsity. The 90\% sparsity model is iteratively pruned from the 80\% sparsity.  The accuracy of the dense model is reported in~\cite{wu2016gnmt}

\subsubsection{ResNet-50 models on image classification}

ResNet-50 models~\cite{he2016deep} are widely used to compare different sparsification methods for convolutional neural networks. We prune all layers to the specified sparsity except for the first convolutional layer, which has larger impact on model accuracy when pruned. 

All sparse models are obtained using the prune-from-dense methodology to remove weights with the least magnitude. We use SGD optimizer with cosine learning rate decay. The initial learning rate is 2.048 and the batch-size is 2048. The total number of finetuning epochs is 250. We train and evaluate on the ImageNet dataset. 

The second rows in  Table~\ref{tab:accuracy-vs-sparsity} lists the top-1 accuracy of the sparse ResNet-50 models with different sparse patterns and sparsity levels. The model is one-shot pruned from a dense model to 60\% sparsity. The 80\% and 90\% sparsity models are iteratively pruned from the 60\% and 80\% sparsity, respectively.

\subsubsection{Jasper on speech recognition}
Jasper~\cite{Li2019JasperAE} is an acoustic speech recognition model consisting of over 200 layers of 1-D convolutional layers with residual connections. We train and evaluate on LibriSpeech dataset with 960 hours of training data. 

All sparse models are obtained using the prune-from-dense methodology to remove weights with the least magnitude. In this task, we compare the weights for {\em all} layers in the Jasper model and then remove them with the least magnitude to obtain a sparse model for a specific sparsity. We use NovoGrad optimizer with an initial learning rate of 0.015 for 400 epochs. The batch size is 64.

The third row in  Table~\ref{tab:accuracy-vs-sparsity} lists the word-error-rate of the sparse Jasper models with different sparse patterns and sparsity levels. The model is one-shot pruned to 77.8\% sparsity. The 83\% and 88.5\% sparsity models are iteratively pruned from the 77.8\% and 83\% sparsity, respectively.

\subsection{Runtime Performance Simulation Setup}
We choose GEM5~\cite{binkert2011gem5} simulator to model the kernel runtime performance of the GS and block patterns. Since the performance is evaluated via a simulator, there is no run-to-run variations and the exact kernel run time can be obtained.

The computer system is illustrated in Figure~2. We choose O3CPU as the processor, which is an eight issue out-of-order CPU model loosely based on Alpha 21264 microarchitecture. We choose ARM instruction set with scalable vector extension (SVE). We use GCC 10.0 to cross compile the kernels to ARM instructions. SVE naturally supports 32- and 64-bit gather/scatter instructions. Since most deep learning models do not require such high resolutions, we implement 16-bit gather/scatter instructions and use them in the kernels. 
The system contains a 64KB L1 cache with two cycle access latency, a 1MB  L2 cache with 20 cycle access latency, and a DDR3 system memory. We implement a 64KB TCM and gather/scatter engine as described in Section~3 with three-cycle access latency without bank conflict, and an extra cycle for every non-resolving bank conflict. The L1 cache comes with a tag prefetcher, which automatically fetch the next four cache lines once a cache access is encountered. The L2 cache comes with a tag prefetcher with block prefetch capability, which means an instruction can specify to prefetch a predetermined memory block. In all performance comparisons, the activations are kept in the TCM and the weights flow through the L1/L2 caches. The activations are partitioned if they are larger than the TCM size.

In all experiments, the weights and activations are saved in 16-bit floating point (FP) values. The values are converted to 32-bit FP values before computation. Accumulation of the partial results are kept in 32-bit FP values as well. The final results are converted back to 16-bit FP values before storing back to memory.

\end{document}